\documentclass[lettersize,journal]{IEEEtran}
\usepackage{amsmath,amsfonts}
\usepackage{algorithmic}
\usepackage{algorithm}
\usepackage{array}
\usepackage{textcomp}
\usepackage{stfloats}
\usepackage{url}
\usepackage{verbatim}
\usepackage{graphicx}
\usepackage{cite}
\usepackage{subfigure}
\usepackage{graphicx}
\usepackage{multirow}
\usepackage{booktabs}
\usepackage{bigstrut}
\usepackage{algorithm}
\usepackage{algorithmic}
\usepackage[table]{xcolor}
\usepackage{balance}
\usepackage{tabularx}
\usepackage{makecell}
\usepackage{threeparttable}

\usepackage{hyperref}
\hypersetup{pdfstartview=FitH,
            colorlinks=true,
            linkcolor=red,
            anchorcolor=blue,
            citecolor=green
            }

\newcounter{RNum}
\renewcommand{\theRNum}{\arabic{RNum}}
\newcommand{\Remark}{\noindent\textit{\textbf{Remark}~\refstepcounter{RNum}\textbf{\theRNum}}: }

\begin{document}

\title{Mutual-Learning Knowledge Distillation for Nighttime UAV Tracking}

\author{Yufeng Liu$^{1}$*
\thanks{
*Corresponding author
}
\thanks{$^{1}$Y. Liu is with the School of Mechanical Engineering, Tongji University, Shanghai 201804, China.
\itshape{Email: 2050520@tongji.edu.cn}
}
}

\markboth{Journal of \LaTeX\ Class Files,~Vol.~14, No.~8, August~2023}%
{Shell \MakeLowercase{\textit{et al.}}: A Sample Article Using IEEEtran.cls for IEEE Journals}

\IEEEpubid{0000--0000/00\$00.00~\copyright~2023 IEEE}

\maketitle

\begin{abstract}
Nighttime unmanned aerial vehicle (UAV) tracking has been facilitated with indispensable plug-and-play low-light enhancers.
However, the introduction of low-light enhancers increases the extra computational burden for the UAV, significantly hindering the development of real-time UAV applications.
Meanwhile, these state-of-the-art (SOTA) enhancers lack tight coupling with the advanced daytime UAV tracking approach.
To solve the above issues, this work proposes a novel mutual-learning knowledge distillation framework for nighttime UAV tracking, \textit{i.e.}, MLKD.
This framework is constructed to learn a compact and fast nighttime tracker via knowledge transferring from the teacher and knowledge sharing among various students.
Specifically, an advanced teacher based on a SOTA enhancer and a superior tracking backbone is adopted for guiding the student based only on the tight coupling-aware tracking backbone to directly extract nighttime object features.
To address the biased learning of a single student, diverse lightweight students with different distillation methods are constructed to focus on various aspects of the teacher's knowledge.
Moreover, an innovative mutual-learning room is designed to elect the superior student candidate to assist the remaining students frame-by-frame in the training phase.
Furthermore, the final best student, \textit{i.e.}, MLKD-Track, is selected through the testing dataset. 
Extensive experiments demonstrate the effectiveness and superiority of MLKD and MLKD-Track.
The practicality of the MLKD-Track is verified in real-world tests with different challenging situations.
The code is available at \href{https://github.com/lyfeng001/MLKD}{https://github.com/lyfeng001/MLKD}.

\end{abstract}

\begin{IEEEkeywords}
Unmanned aerial vehicle (UAV), nighttime UAV tracking, mutual-learning knowledge distillation, tight coupling-aware student model, superior tracking performance.
\end{IEEEkeywords}

\section{Introduction}
\IEEEPARstart{V}{isual} object tracking is crucial for intelligent unmanned aerial vehicle (UAV) applications such as monitoring, patrolling, and surveying~\cite{fu2023siamese,Ding2021DesignSA, Li2020AutotrackTH}.
Given an initial object position, trackers are expected to estimate the subsequent location of a tracked object.
With high-quality daytime training datasets~\cite{fan2021lasot,8922619}, the state-of-the-art (SOTA) UAV trackers~\cite{siamrpn++, hift, fu2023continuity} perform considerably in favorable illumination scenes.
However, these SOTA trackers are severely degraded in nighttime UAV tracking situations due to the weaknesses of nighttime images compared with daytime images, including low brightness, low contrast, appearance distortion, and low signal-to-noise ratio.
Therefore, the backbone trained with daytime images loses its effectiveness in extracting nighttime object features, leading to unsatisfying tracking failures at night.

To sustain trackers' remarkable performance from daytime to nighttime conditions, a promising solution is to incorporate a low-light enhancer~\cite{darklighter, sct, highlightnet,ruas,enlightengan,zhang2021learning,li2021learning} before the advanced daytime tracker with a plug-and-play strategy.
Through this preprocessing step, the nighttime images are enhanced with favorable illumination and ample details.
The daytime SOTA trackers can achieve impressive results using these enhanced nighttime images.
However, constrained by limited computational resources, algorithms with high computational demands become impractical for UAVs.
The extra computation complexity of low-light enhancers impedes the development of real-time nighttime UAV applications. 
Moreover, loose coupling between the enhancer and the SOTA tracker limits nighttime UAV tracking performance.
\textit{How to realize a robust nighttime UAV tracking performance without using the plug-and-play low-light enhancers while maintaining a fast speed is a pressing challenge.}\IEEEpubidadjcol

In literature, knowledge distillation is frequently employed to transfer specialized knowledge from complex models to more lightweight counterparts.
This method has been widely used in computer vision tasks, including image classification~\cite{fitnet}, object detection~\cite{ld}, image segmentation~\cite{liu2019structured}, \textit{etc}.
During the knowledge distillation process, the trained student model usually exhibits strong generalization ability since the distilled knowledge contains more specific details~\cite{distillorigin}.
As a result, the distilled student model can exceed the accuracy of the ordinary student model, demonstrating performance on par with that of the teacher model.
Although knowledge distillation has demonstrated excellent performance in model compression and inference acceleration, using a single student with a fixed distillation method leads to the student model focusing on a specific aspect, resulting in biased learning.
To acquire knowledge from various perspectives, different distillation methods can be utilized to obtain diverse lightweight student models.
Sharing knowledge among all the students has the potential to avoid biased learning caused by a single student.
Therefore, it becomes possible to create a more robust model for nighttime UAV tracking by sharing knowledge among students to integrate their strengths.

This work proposes a mutual-learning knowledge distillation framework (MLKD) to learn a fast and compact nighttime UAV tracker. 
The advanced teacher is based on a SOTA enhancer with a superior tracking backbone.
The students based on only the tracking backbone learn from the teacher to directly extract nighttime object features.
Different distillation methods are employed to gain diverse lightweight students with different emphases.
A novel mutual-learning framework is established to dynamically choose the best student to assist other students frame-by-frame during the training phase.
The best student model after training, \textit{i.e.}, MLKD-Track, is selected for inference.
Having the same number of parameters as an advanced baseline tracker, MLKD-Track can still possess superior tracking performance in nighttime scenes.

The contributions of this work lie in four-fold:
\begin{figure*}[tbp]
\centering
\includegraphics[width=\linewidth,scale=1.00]{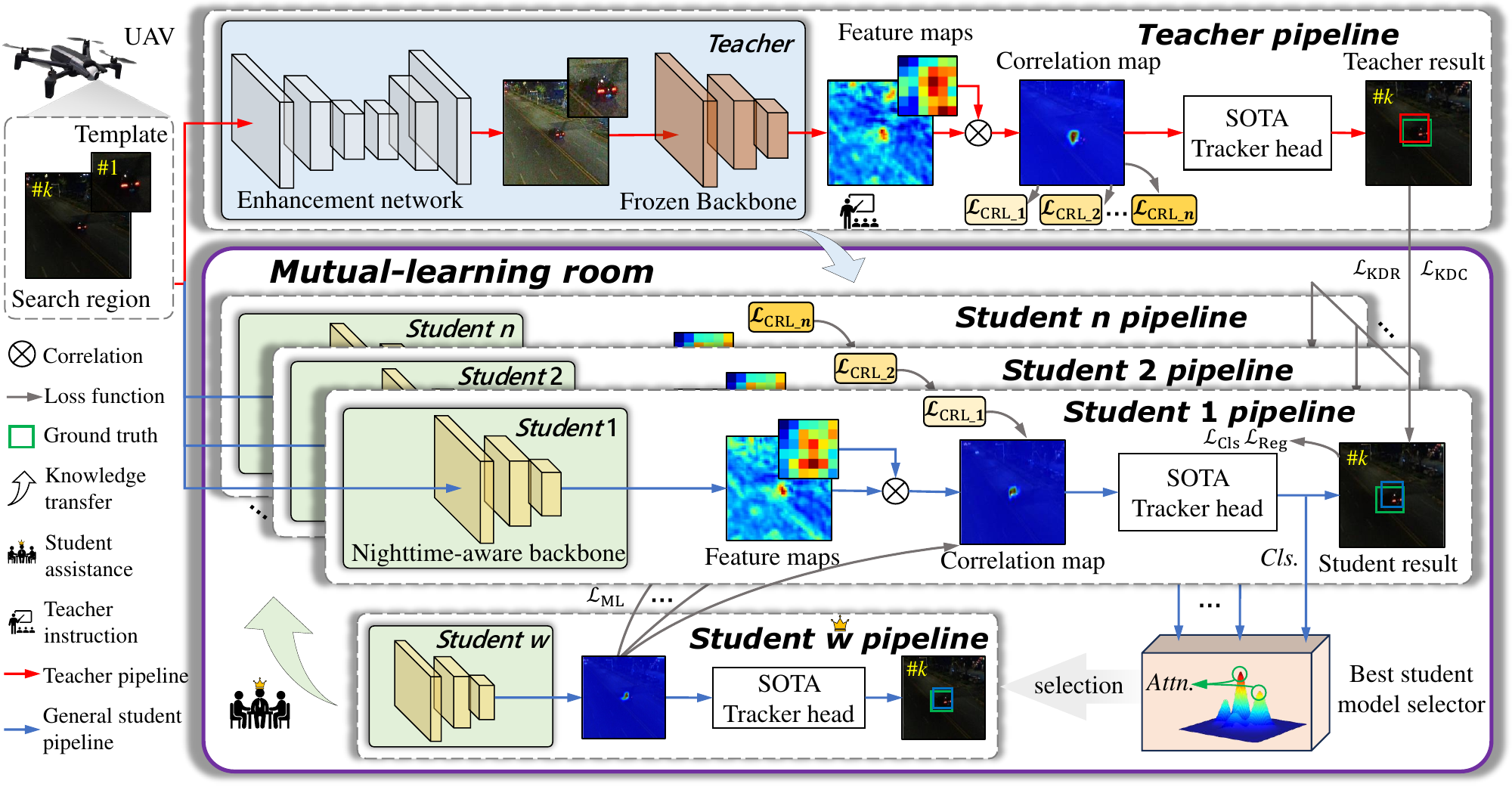}
\caption{Overview of the presented mutual-learning knowledge distillation framework. It contains an advanced teacher model and multiple student models. Each student model learns from the teacher model with a specific emphasis through a correlation loss function. In the training phase, the best student is selected in each frame to assist the other student models with the mutual learning loss. In the testing phase, the selected final best student model can extract features from the low-light images directly.}
\label{fig:QD}
\end{figure*}
\begin{itemize}
    \item A novel mutual-learning knowledge distillation framework (MLKD) has been developed to accomplish robust nighttime UAV tracking with a lightweight tight coupling-aware model. 
    MLKD removes the additional computational burden and achieves the tight coupling between the SOTA enhancer and the advanced daytime tracking approach.
    
    \item Multiple lightweight students with different distillation methods are designed to learn from the teacher with different focuses.

    \item A new mutual-learning room is constructed to facilitate mutual complementation of strengths among students, achieving comprehensive learning from the teacher and superior performance in nighttime UAV tracking.

    \item The superiority of the proposed framework has been proved with extensive evaluation experiments. Compared with the SOTA tracking methods, the advantages of MLKD-Track can be testified. Real-world tests verify the practicality of MLKD-Track facing various situations.
\end{itemize}

\section{Related Works}
\subsection{UAV Object Tracking}
Visual object tracking has been widely used on intelligent robots such as UAVs~\cite{9991169,zheng2021mutation,siamfc,siamapn++,siamapn,siamrpn,zuo2022end}. 
With the continuous development of intelligent robots, the Siamese network-based methods gradually become the current mainstream of UAV tracking due to their superior success rate and precision~\cite{fu2023siamese}.
B. Li \emph{et al.}~\cite{siamrpn} introduce the region proposal network (RPN) and SiamRPN++~\cite{siamrpn++} further proposes a new sampling strategy to break the spatial invariance restriction and constructs a deep correlation method.
To increase the robustness and generalization of different objects with different challenges, SiamAPN~\cite{siamapn} designs an innovative anchor proposal method.
To improve feature representation, SiamAPN++~\cite{siamapn++} exploits the attention mechanism.
TCTrack~\cite{cao2022tctrack} constructs a comprehensive framework for UAV tracking to take full advantage of the temporal contexts in the network.
Despite remarkable performance in the daytime, these trackers struggle to predict the object location precisely in the nighttime scenes due to the substantial difference between daytime and nighttime images.

\subsection{Low-Light UAV Tracking}
As UAV tracking becomes more and more widely used, many studies~\cite{darklighter,sct,highlightnet,udat,yao2023sam} on UAV object tracking in nighttime environments improve the UAV tracking performance.
DarkLighter~\cite{darklighter} enhances low-light images to improve UAV tracking performance in nighttime scenes by utilizing unsupervised training while accounting for noise.
To achieve semantic-level low-light enhancement, J. Ye \emph{et al.}~\cite{sct} construct a spatial-channel Transformer-based low-light enhancer.
HighlightNet~\cite{highlightnet} adapts to illumination variation and excavates the potential object for low-light UAV tracking.
Although most low-light UAV tracking methods employ relatively straightforward strategies, they are all extra additions to the basic UAV tracking structure.
Additionally, an enhanced image is not completely identical to a daylight image due to the introduction of unfavorable disturbance during the enhancement process.
Some original details in that image are lost in this process.
Therefore, a lack of tight coupling between the enhancer and the SOTA tracker restrains the performance of nighttime UAV tracking.
It is vitally necessary to develop a method that enables direct feature extraction from low-light images utilizing a lightweight backbone.

\subsection{Knowledge Distillation}
Knowledge distillation~\cite{distillorigin} is initially presented to achieve a model-compressing effect by using a larger model to guide the training of a smaller model.
The generalization ability of the small models is improved by the use of soft labels in the computation of the loss function.
With the expansion of the applications, knowledge distillation is deployed in image classification~\cite{fitnet}, object detection~\cite{ld}, image segmentation~\cite{liu2019structured}, \emph{etc}.
Maintaining a performance near a complex model, the smaller model can have a more rapid speed.
FitNet~\cite{fitnet} proposes a feature-level distillation loss function to obtain a better distillation effect.
Focusing on transferring structured pixel-to-pixel and pixel-to-region relations among the whole images, CIRKD~\cite{yang2022cross} applies knowledge distillation in the image segmentation field.
Z. Zheng \emph{et al.}~\cite{ld} combine classification and localization knowledge and obtain them simultaneously at the detector head.
However, these methods only involve a student model with a fixed distillation method during the training process, restricting comprehensive knowledge transferring, particularly in complex nighttime UAV tracking scenarios.

\section{Methodology}
This work proposes a mutual-learning knowledge distillation framework for nighttime UAV tracking.
The framework is composed of a teacher model and multiple student models, as illustrated in Fig.~\ref{fig:QD}.
In this knowledge distillation framework, the knowledge transferring is divided into two parts, \textit{i.e.}, 1) the knowledge transfer from the teacher to the students, and 2) the knowledge transfer among the student models.

\subsection{Structure of Teacher and Student Models}
This framework is used to distill any kind of low-light enhancers with baseline trackers.
For UAV tracking, a lightweight model is required to adapt to the UAV tracking task.
Thereby, the baseline tracker is based on the SiamRPN++~\cite{siamrpn++}, which uses AlexNet~\cite{alexnet} as the backbone.
Meanwhile, a SOTA low-light enhancer~\cite{darklighter} is adopted to preprocess the nighttime images.
Specifically, the teacher model is composed of the low-light enhancer and the baseline tracker, while the student models only contain the baseline tracker.

\Remark
This framework can be used to process different low-light enhancers plus baseline trackers. As a lightweight backbone, AlexNet is a suitable backbone for UAV tracking. 

\subsection{Tracking-Oriented Knowledge Distillation}
For the knowledge transfer from the teacher to the student models, the loss functions are elaborately designed. 
In addition to the original classification and regression losses, several knowledge distillation losses are proposed to facilitate the knowledge-transferring process.
In terms of classification, the tracker head of all the models can classify in the proposal region of the feature maps, outputting the logit values belonging to the foreground and background.
The logit values of the teacher model are softened as the soft label, while the logits output by the student model are also softened.
There is a knowledge distillation loss for classification, \textit{i.e.}, $\mathcal{L}_{\mathrm{KDC}}$:
\begin{equation}\label{eqn:1}
\mathcal{L}_{\mathrm{KDC}}=\mathcal{L}_{\mathrm{KL}}(\mathcal{S}(c_S,\tau),\mathcal{S}(c_T,\tau)) \quad, 
\end{equation}
where $c_S$ is the classification logit values of the student model, and $c_T$ is that of the teacher model.
Meanwhile, $\mathcal{L}_{\mathrm{KL}}$ represents the KL-Divergence loss of its following varieties.
The $\mathcal{S}(\cdot,\tau)$ represents a softmax method with a distillation temperature $\tau$, \textit{i.e.}:
\begin{equation}\label{eqn:2}
\mathcal{S}(\cdot,\tau)=\mathrm{SoftMax}(\cdot/\tau)\quad.
\end{equation}

Referring to the positional distillation loss in the LD knowledge distillation method, the regression label can also engage in the backward process as a part of the soft loss~\cite{ld}.
Concretely. the regression label and the predicted labels are all softened and combined into a soft loss, \textit{i.e.}, $\mathcal{L}_{\mathrm{KDR}}$:
\begin{equation}\label{eqn:3}
\mathcal{L}_{\mathrm{KDR}}=\mathcal{L}_{\mathrm{KL}}(\mathcal{S}(r_S,\tau),\mathcal{S}(r_T,\tau))\quad , 
\end{equation}
where $r_S$ is the regression logit values of the student model, and $r_T$ is that of the teacher model.

In order to learn from the teacher model in different aspects, different distillation methods need to be designed.
In this framework. the enhancer leads to a significant difference in feature maps between the teacher and student models.
It is necessary to avoid the direct approximation of the student output to the teacher output after the middle layers, as this ignores the image changing after the enhancer and potentially mislead the students.
Instead, the correlation maps are chosen for changing the distillation methods since the large response values on them correspond to the possible target location, which is invariant.
Therefore, a loss function corresponding to the correlation maps \textit{i.e.}, $\mathcal{L}_{\mathrm{CRL}}$, can be proposed as:
\begin{equation}\label{eqn:4}
\mathcal{L}_{\mathrm{CRL}}={\mathcal{F}_{\it{n}}}(\mathcal{C}_{S\it{n}},\mathcal{C}_T) \quad, 
\end{equation}
where the $\mathcal{C}_{S\it{n}}$ represents the correlation maps of the student model $\mathcal{\it{n}}$, and the $\mathcal{C}_T$ represents that of the teacher model.
The $\mathcal{F_{\it{n}}}$ represents to a specific loss function of the student model $\mathcal{\it{n}}$.
The distillation methods can be changed through correlation loss function alternation.
For each student, the correlation loss functions are designed to focus on different aspects, leading students to learn from diverse emphases.

\Remark
Since different correlation loss functions are designed, diverse students can approximate the teacher from different directions to gain different unique knowledge. 

\begin{figure}[t]
\centering
\includegraphics[width=\linewidth,scale=1.00]{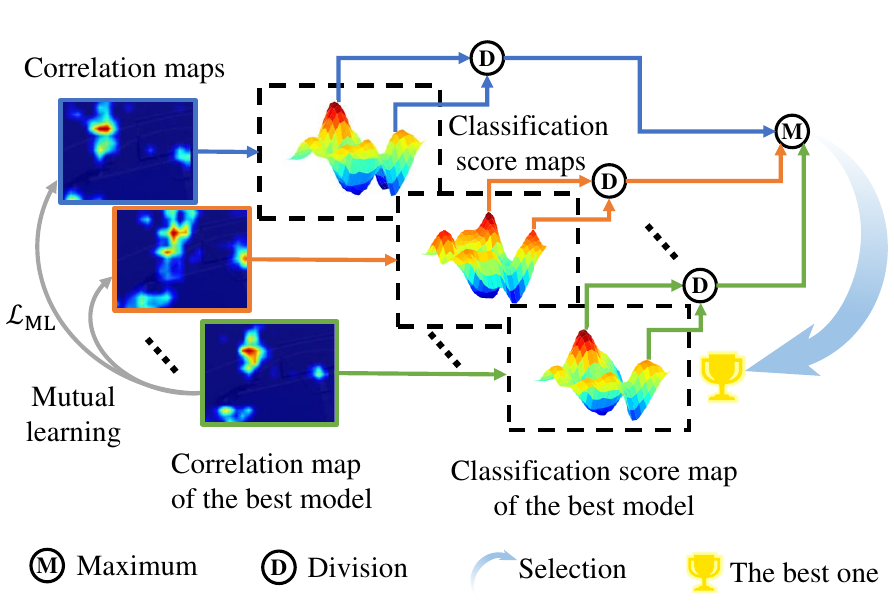}
\caption{The details of the mutual learning. Persuasive values are calculated by the per-pixel classification score maps of different student models. The maximum persuasive value corresponds to the best student model. The best student model assists the other student models with a correlation loss function. (Image frames are from UAVDark135~\cite{adtrack}.)}
\label{fig:2}
\end{figure}
\subsection{Mutual-Learning Room}
In the mutual-learning room, distinct student models are generated with different loss functions.
The generalization ability of the student models is improved and the student models are able to extract object features directly from the nighttime images.
However, each student model has strengths in specific aspects, and can not learn comprehensively from the teacher model.
Therefore, it limits the performance of the student models.
By sharing the useful knowledge acquired by the student models, it can compensate for the biased learning of the independent single-student knowledge distillation.

To accomplish the sharing of knowledge obtained by the student models, the best student model of one specific frame is selected.
Concretely, the classification score maps output by the student models are used to determine if a pixel position corresponds to the target object.
The local maximum points indicate the likely center of the target box and the absolute maximum value point locates at the center of the target box.
However, multiple local maximum value points can be generated in one classification score map.
In UAV tracking, the number and the value of the peaks reflect the confidence of the model's tracking results.
Therefore, a smaller number of peaks and a relatively larger absolute maximum value indicate greater confidence, and the model should be more persuasive.
As shown in Fig.~\ref{fig:2}, to determine the persuasive value of each model, all local maximum values in the classification score maps are selected.
The ratio of the maximum value to the second largest value is taken as the persuasive value.
Moreover, the model with the largest persuasive value is chosen as the best model of the current frame.

For each frame, the best student model transfers its knowledge to the other student models through another correlation loss.
Meanwhile, a coefficient is designed to take the number of student models into account:
\begin{equation}\label{eqn:5}
\mathcal{L}_{\mathrm{ML}}=\frac{1}{u}\sum_{i=1}^{u-1}({\mathrm{MSE}}(\hat{\mathcal{C}_S},\mathcal{C}_{S\it{i}})) \quad,
\end{equation}
where the $\hat{\mathcal{C}_S}$ represents correlation maps of the best student model and $u$ is the number of the student models.
The parameters of the best student model should be frozen in backward propagation of this correlation loss.

\begin{figure}[t]
\centering
\includegraphics[width=\linewidth,scale=1.00]{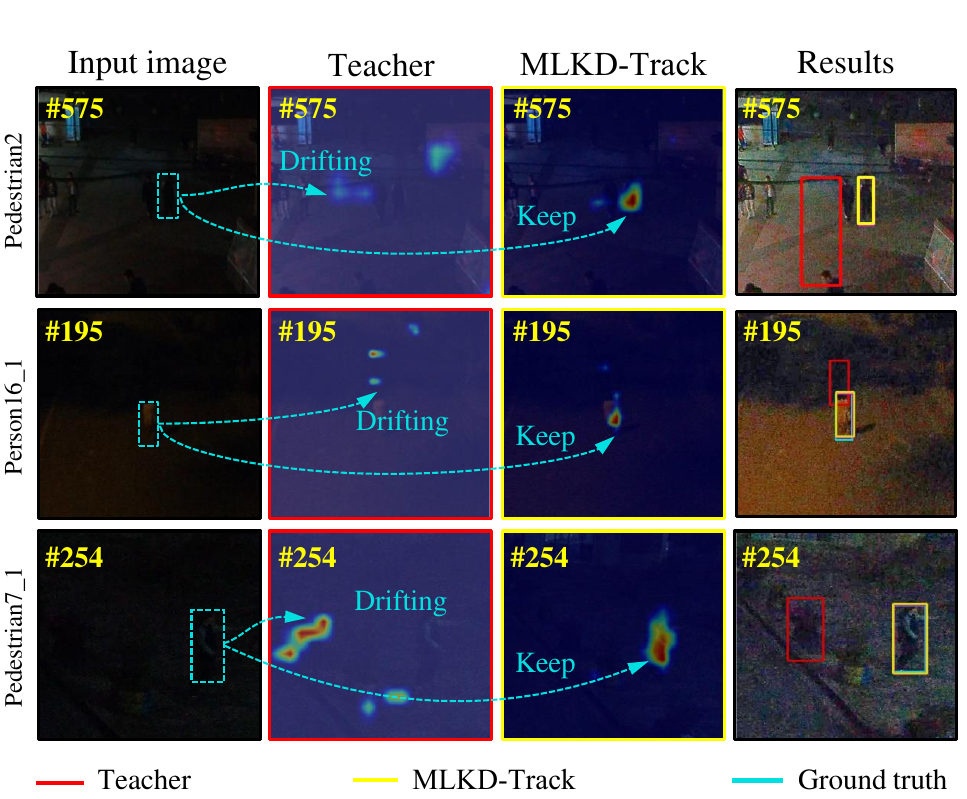}
\caption{Comparison of the similarity heatmaps between MLKD-Track and the teacher model. The proposed MLKD-Track shows better performance in some specific scenes. To show the tracking result of different trackers better, the tracking results have been enhanced by DarkLighter. (Image frames are from UAVDark135~\cite{adtrack}.)} \label{fig:3}
\end{figure}

In conclusion, the overall loss function can be ascribed as:
\begin{equation}\label{eqn:6}
\begin{split}
\mathcal{L}=\lambda_{1}(\mathcal{L}_{\mathrm{Cls}}+\mathcal{L}_{\mathrm{Reg}})+&\lambda_{2}(\mathcal{L}_{\mathrm{KDC}}+\mathcal{L}_{\mathrm{KDR}})\\+&\lambda_{3}\mathcal{L}_{\mathrm{CRL}}+\lambda_{4}\mathcal{L}_{\mathrm{ML}}\quad, 
\end{split}
\end{equation}
As shown in Fig.~\ref{fig:3}, the final best student model, \textit{i.e.}, MLKD-Track can acquire knowledge beyond the teacher, performing even better than the teacher in some circumstances.

\Remark
Using the ratio of the first two maximum values as a persuasive value can choose the best student model frame by frame to assist the other student models in learning its strengths. 
Therefore, all the student models can learn more comprehensive knowledge than the student models without mutual learning. 

\section{Experiment}
In this section, the student models are trained through the mutual-learning knowledge distillation framework, and the final best student model, \textit{i.e.}, MLKD-Track is selected.
The MLKD-Track is compared with other SOTA trackers.
The superiority of MLKD-Track and the MLKD can be testified through these experiments. 
The ablation study and the real-world tests verify the feasibility and practicality of MLKD-Track.

\subsection{Evaluation Metrics}
To test the feasibility and superiority of the mutual-learning knowledge distillation framework, the UAVDark135 benchmark~\cite{adtrack} is used as a testing dataset.
All models are tested under the same condition.
The evaluation complies with the one-pass evaluation (OPE)~\cite{OPE}.
For the models, the success rate, normalized precision, and precision are compared to judge their strengths and weaknesses.
The success rate is calculated by the intersection over the union of the predicted bounding box and the ground truth.
The precision is the Euclidean distance error between the center of the output box and the ground truth.

\subsection{Implementation Details}
During the training and testing process, the experiments are conducted on four NVIDIA A100 GPUs.
The initial learning rate is 0.005 and the batchsize is 64. 
In order to achieve better training results, the coefficient of knowledge distillation loss $\lambda_{2}$, the coefficient of correlation graph loss $\lambda_{3}$, the coefficient of the mutual-learning loss $\lambda_{4}$, and the coefficient of classification and regression loss $\lambda_{1}$ is chosen to be 1000, 0.2, 20, and 1.
The NAT2021~\cite{udat} and DarkTrack2021~\cite{sct} are used as the training datasets.

The teacher model is based on the SOTA enhancer~\cite{darklighter} and the advanced tracker~\cite{siamrpn++}, and the student model is only based on the advanced tracker~\cite{siamrpn++}. 
The parameters of the teacher model are frozen in the training phase, while the parameters of the student model are also frozen except for the tracking backbone.
The mutual-learning knowledge distillation framework contains $n$ student models.
In the experiment, $n$ is chosen to be 3 in order to illustrate the framework with a small number of student models.
The number of student models can vary in different applications.

As discussed in the previous section, each student model is associated with its own correlation loss function. 
The first correlation loss function utilizes the L2 loss to approximate the distance of numerical values between the two correlation maps:
\begin{equation}\label{eqn:7}
\mathcal{L}_{\mathrm{CRL}\_\mathrm{L2}}={\mathrm{MSE}}(\mathcal{C}_S,\mathcal{C}_T) \quad, 
\end{equation}
where MSE represents the mean square error. 
$\mathcal{C}_S$ represents the correlation maps of the student model, and $\mathcal{C}_T$ means that of the teacher model.

The second type of correlation loss function is spatial consistency loss, where the correlation maps are flattened and then SoftMax is applied to them.
Furthermore, the disparity between the value at each pixel point and the four adjacent neighboring values is calculated. 
The cumulative difference value of the two processed correlation maps is compared to obtain the correlation loss function, denoted as:
\begin{equation}\label{eqn:8}
\mathcal{L}_{\mathrm{CRL}\_\mathrm{S}}=\frac{1}{m}\sum_{i=1}^{m}\mathrm{(\mathcal{N(\it{i})}_{\mathrm{S}}-\mathcal{N(\it{i})}_{\mathrm{T}})^2} \quad, 
\end{equation}
where $\it{m}$ is the number of pixels in the correlation maps of one student model.
$\mathcal{N(\it{i})}$ represents the total of the four gaps between a pixel and each of its neighbors, \textit{i.e.}:
\begin{equation}\label{eqn:9}
\mathcal{N(\it{i})}=\sum_{\it{u}=1}^{4}(\mathcal{P}_{\it{i}}-\mathcal{P}_{\it{iu}}^n) \quad, 
\end{equation}
where $\mathcal{P}_{\it{i}}$ represents one of the pixels in the correlation maps, and $\mathcal{P}_{\it{iu}}^n$ represents one of the four neighbors of the pixel ahead~\cite{dce}.
With the spatial consistency loss, differences in the changing trend within each correlation map can be measured at the pixel level.
Therefore, it can bring the student model output closer to the teacher model in the pixel-changing aspect.

Another loss function is concerned with a response threshold, in which the correlation maps are also flattened first.
The average value of each vector is computed and used as the response threshold.
Element in the vector is set to 1 when it is greater than the threshold while being set to 0 when it is smaller than the threshold.
\begin{figure*}[htbp]
\centering
\subfigure {
\includegraphics[width=0.31\linewidth,scale=1.00]{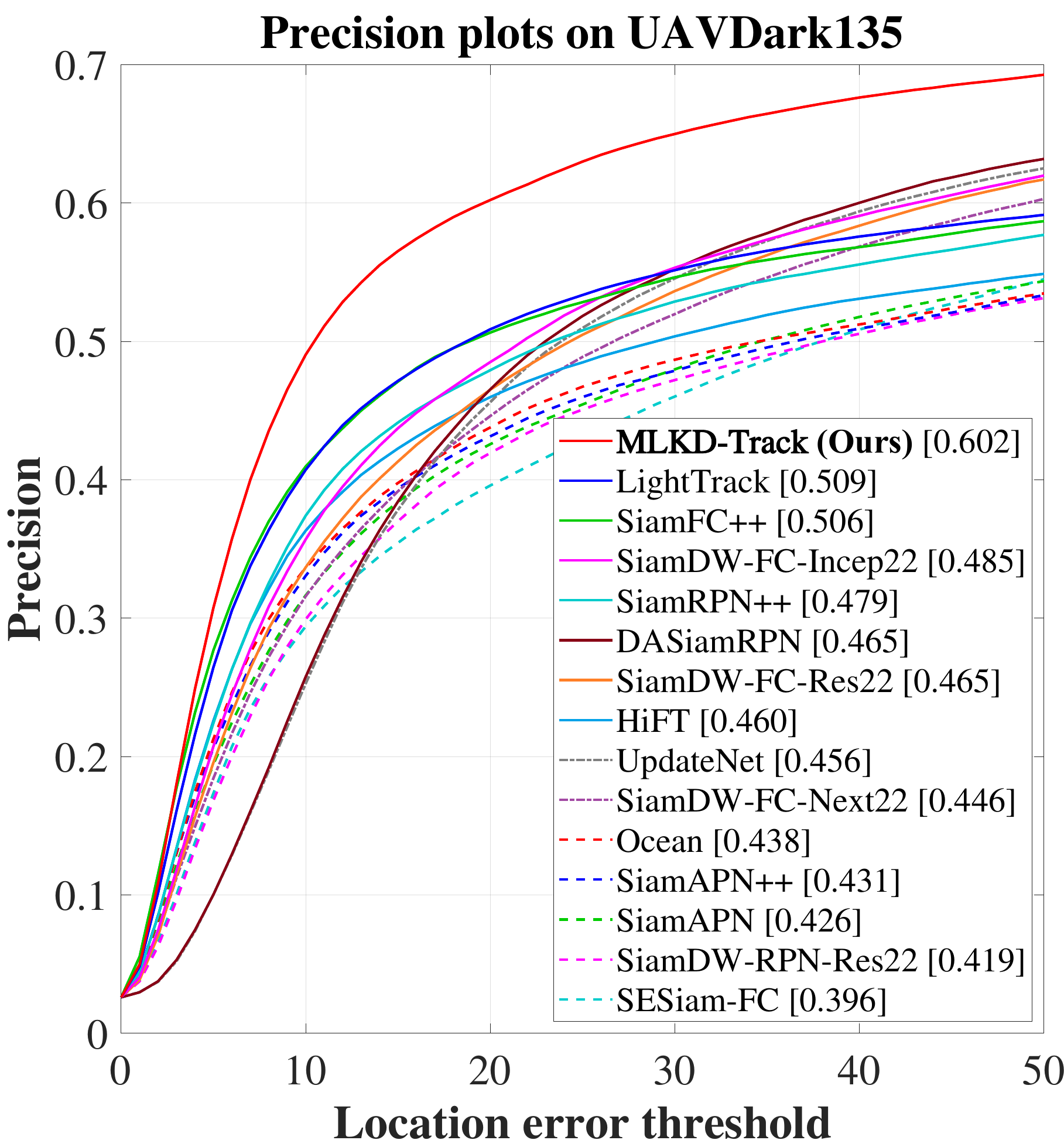}
}
\subfigure {
\includegraphics[width=0.31\linewidth,scale=1.00]{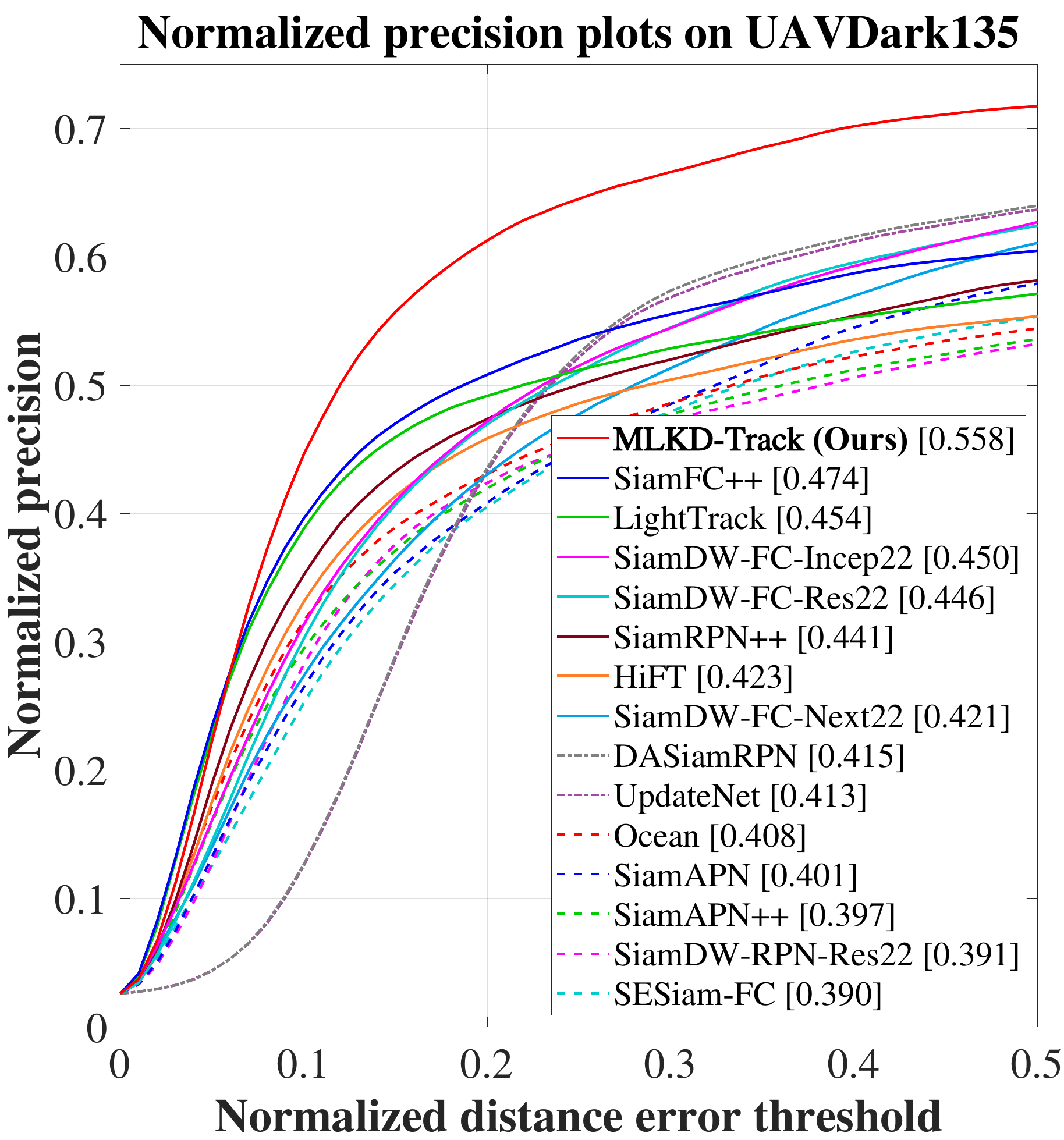}
}
\subfigure {
\includegraphics[width=0.31\linewidth,scale=1.00]{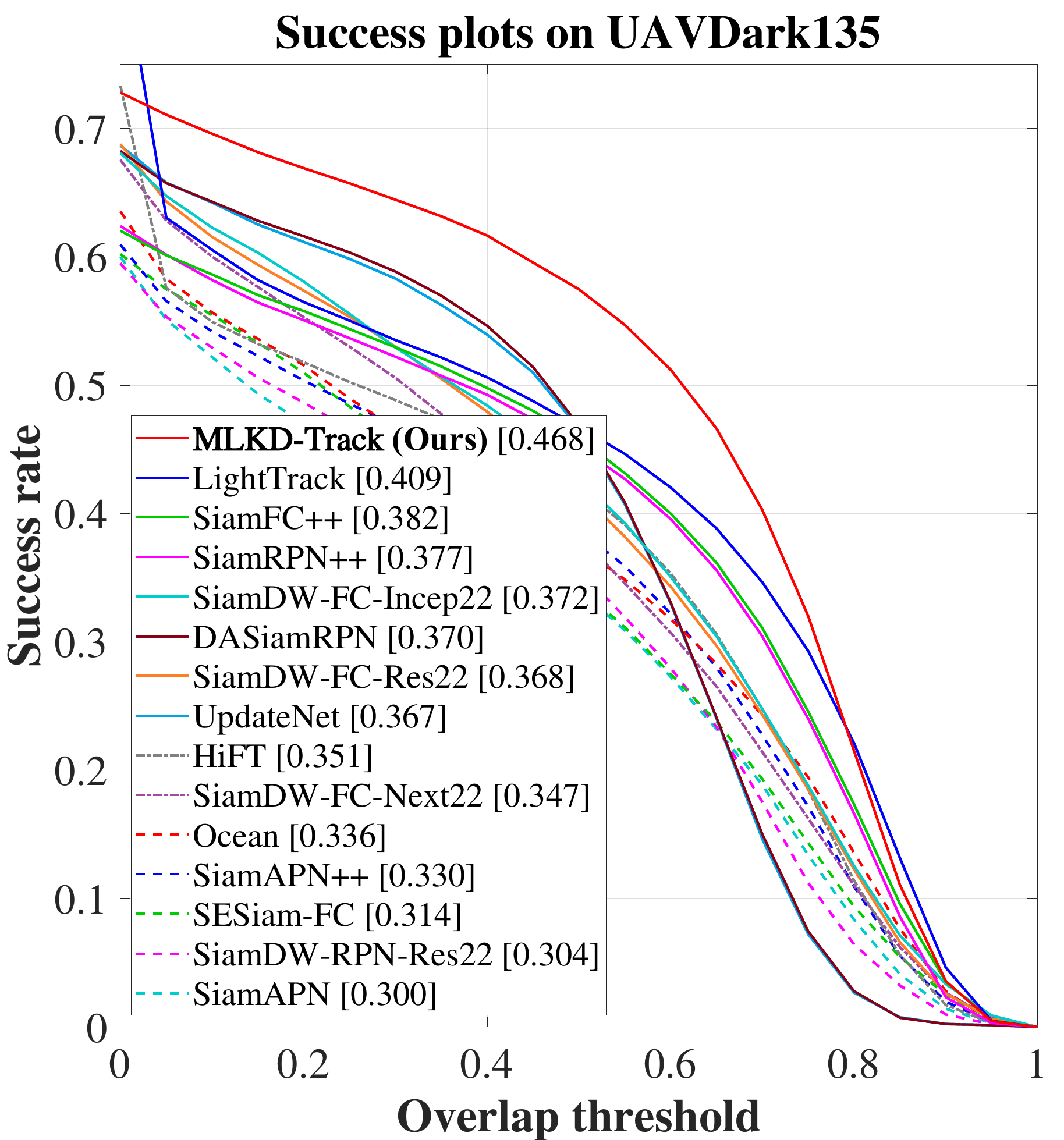}
}
\caption{Precision and success plots of MLKD-Track and other SOTA trackers~\cite{hift,siamrpn++,yan2021lighttrack,xu2020siamfc++,zhang2019deeper,zhang2020ocean,zhu2018distractor,li2021learning,siamapn,siamapn++,sosnovik2021scale} on the UAVDark135 benchmark~\cite{adtrack}. With the MLKD framework, MLKD-Track performs the best while keeping an uncomplicated structure that is more suitable for UAV tracking.}
\label{fig:rainbow}
\vspace{-4pt}
\end{figure*}
\begin{table*}[htbp]
  \centering
  \setlength{\tabcolsep}{5pt}
  \caption{Attribute-based evaluation of the MLKD-Track and other 14 SOTA trackers on UAVDark135~\cite{adtrack}. The best three performance are respectively highlighted with \textcolor[rgb]{1,0,0}{red}, \textcolor[rgb]{ 0,  .69,  .314}{green}, and \textcolor[rgb]{0,  .439,  .753}{blue}, $\Delta$ denotes the improvement in comparison with the second best tracker.}
    \begin{tabular}{lccccccccccccccc}
    \toprule
    Attributes & \multicolumn{3}{c}{Fast motion} & \multicolumn{3}{c}{Illumination variation} & \multicolumn{3}{c}{Low resolution} & \multicolumn{3}{c}{Occlusion} & \multicolumn{3}{c}{Viewpoint change} \\
    \midrule
    Trackers & Prec. & Norm. & Succ. & Prec. & Norm. & Succ. & Prec. & Norm. & Succ. & Prec. & Norm. & Succ. & Prec. & Norm. & Succ. \\
    \midrule
    LightTrack & \textcolor[rgb]{0,  .69,  .314}{0.471} & \textcolor[rgb]{ 0,  .439,  .753}{0.433} & \textcolor[rgb]{ 0,  .69,  .314}{0.384} & \textcolor[rgb]{ 0,  .439,  .753}{0.483} & \textcolor[rgb]{ 0,  .439,  .753}{0.430} & \textcolor[rgb]{ 0,  .69,  .314}{0.395} & \textcolor[rgb]{ 0,  .439,  .753}{0.507} & \textcolor[rgb]{ 0,  .439,  .753}{0.433} & \textcolor[rgb]{ 0,  .69,  .314}{0.382} & {0.447} & {0.406} & \textcolor[rgb]{ 0,  .439,  .753}{0.368} & \textcolor[rgb]{ 0,  .439,  .753}{0.484} & \textcolor[rgb]{ 0,  .439,  .753}{0.448} & \textcolor[rgb]{ 0,  .439,  .753}{0.414} \\
    SiamFC++ & \textcolor[rgb]{ 0,  .439,  .753}{0.461} & \textcolor[rgb]{ 0,  .69,  .314}{0.443} & {0.351} & \textcolor[rgb]{ 0,  .69,  .314}{0.503} & \textcolor[rgb]{ 0,  .69,  .314}{0.454} & \textcolor[rgb]{ 0,  .439,  .753}{0.374} & 0.499  & \textcolor[rgb]{ 0,  .69,  .314}{0.441} & 0.355  & \textcolor[rgb]{ 0,  .69,  .314}{0.485} & \textcolor[rgb]{ 0,  .69,  .314}{0.459} & {0.364} & \textcolor[rgb]{ 0,  .69,  .314}{0.529} & \textcolor[rgb]{ 0,  .69,  .314}{0.509} & \textcolor[rgb]{ 0,  .69,  .314}{0.419} \\
    SiamDW-FC-Incep22 & 0.441  & 0.424  & 0.347  & 0.464  & 0.422  & 0.365  & \textcolor[rgb]{ 0,  .69,  .314}{0.516} & 0.426  & \textcolor[rgb]{ 0,  .439,  .753}{0.356} & 0.414  & 0.404  & 0.330  & 0.400  & 0.411  & 0.346  \\
    SiamRPN++ & 0.456  & 0.431  & \textcolor[rgb]{ 0,  .439,  .753}{0.368}  & 0.435  & 0.393  & 0.339  & 0.489  & 0.412  & 0.348  & \textcolor[rgb]{ 0,  .439,  .753}{0.467}  & \textcolor[rgb]{ 0,  .439,  .753}{0.437}  & \textcolor[rgb]{ 0,  .69,  .314}{0.377}  & 0.413  & 0.411  & 0.363  \\
    SiamDW-FC-Res22 & 0.410  & 0.410  & 0.336  & 0.432  & 0.405  & 0.340  & 0.467  & 0.405  & 0.326  & 0.379  & 0.386  & 0.318  & 0.348  & 0.386  & 0.332  \\
    HiFT & 0.428  & 0.405  & 0.327  & 0.420  & 0.379  & 0.326  & 0.492  & 0.425  & 0.351  & 0.434  & 0.402  & 0.326  & 0.422  & 0.416  & 0.354  \\
    SiamDW-FC-Next22 & 0.403  & 0.398  & 0.322  & 0.409  & 0.375  & 0.316  & 0.470  & 0.395  & 0.318  & 0.354  & 0.358  & 0.295  & 0.368  & 0.386  & 0.326  \\
    DASiamRPN & 0.443  & 0.404  & 0.357  & 0.452  & 0.391  & 0.352  & 0.506  & 0.391  & 0.348  & 0.430  & 0.391  & 0.348  & 0.406  & 0.409  & 0.375  \\
    UpdateNet & 0.433  & 0.402  & 0.355  & 0.441  & 0.388  & 0.350  & 0.493  & 0.383  & 0.343  & 0.426  & 0.392  & 0.348  & 0.407  & 0.413  & 0.378  \\
    Ocean & 0.414  & 0.395  & 0.321  & 0.420  & 0.385  & 0.320  & 0.459  & 0.396  & 0.319  & 0.418  & 0.384  & 0.316  & 0.369  & 0.377  & 0.311  \\
    SiamAPN & 0.400  & 0.384  & 0.282  & 0.402  & 0.376  & 0.279  & 0.452  & 0.387  & 0.290  & 0.412  & 0.400  & 0.298  & 0.438  & 0.435  & 0.330  \\
    SiamAPN++ & 0.403  & 0.380  & 0.314  & 0.405  & 0.367  & 0.309  & 0.464  & 0.395  & 0.324  & 0.415  & 0.390  & 0.328  & 0.425  & 0.410  & 0.351  \\
    SiamDW-RPN-Res22 & 0.385  & 0.376  & 0.291  & 0.394  & 0.360  & 0.283  & 0.394  & 0.349  & 0.271  & 0.410  & 0.391  & 0.305  & 0.362  & 0.357  & 0.281\\
    SESiam-FC & 0.348  & 0.358  & 0.281  & 0.379  & 0.363  & 0.296  & 0.422  & 0.379  & 0.298  & 0.329  & 0.336  & 0.272  & 0.334  & 0.360  & 0.296  \\
    \textbf{MLKD-Track (Ours)} & \textcolor[rgb]{ 1,  0,  0}{0.588} & \textcolor[rgb]{ 1,  0,  0}{0.557} & \textcolor[rgb]{ 1,  0,  0}{0.463} & \textcolor[rgb]{ 1,  0,  0}{0.617} & \textcolor[rgb]{ 1,  0,  0}{0.562} & \textcolor[rgb]{ 1,  0,  0}{0.478} & \textcolor[rgb]{ 1,  0,  0}{0.622} & \textcolor[rgb]{ 1,  0,  0}{0.542} & \textcolor[rgb]{ 1,  0,  0}{0.455} & \textcolor[rgb]{ 1,  0,  0}{0.589} & \textcolor[rgb]{ 1,  0,  0}{0.538} & \textcolor[rgb]{ 1,  0,  0}{0.448} & \textcolor[rgb]{ 1,  0,  0}{0.549} & \textcolor[rgb]{ 1,  0,  0}{0.540} & \textcolor[rgb]{ 1,  0,  0}{0.458} \\
    \midrule
    $\Delta$ (\%) & 24.8  & 25.7  & 20.6  & 22.7  & 23.8  & 21.0  & 20.5  & 22.9  & 19.1  & 21.4  & 17.2  & 21.7  & 3.8  & 6.1  & 9.3  \\
    \bottomrule
    \end{tabular}%
  \label{tab:II}%
  \vspace{-10pt}
\end{table*}%
The difference between these two processed vectors is then measured to obtain the response threshold loss:
\begin{equation}\label{eqn:10}
\mathcal{L}_{\mathrm{CRL}\_\mathrm{R}}=\frac{1}{m}\sum_{i=1}^{m}(\mathcal{B}_{\mathrm{S}\it{i}}-\mathcal{B}_{\mathrm{T}\it{i}}), 
\\ \mathcal{B}_{\it{i}}=\left\{
\begin{array}{rcl}
0,&\mathcal{P}_{\it{i}}<\overline{\mathcal{P}} \\
1,&\mathcal{P}_{\it{i}}>\overline{\mathcal{P}}
\end{array}
\right.,
\end{equation}
This loss function blurs the specific value of the response in the correlation maps, allowing the part of the correlation maps which has larger response values to be approximated.
It is beneficial for optimizing the localization of larger targets and can focus on the edge information of the target objects.

When training without mutual learning, the student with L2 loss is designated as Student 1 model, while the student with spatial consistency loss is designated as Student 2 model, and the student with response threshold loss is designated as Student 3 model.
Meanwhile, other loss functions can be used to construct other student models, learning from the teacher with other emphases.
Three of the correlation loss functions are discussed in this experiment.

When training with mutual learning, the student models are trained simultaneously.
The final student who performs the best on the UAVDark135~\cite{adtrack} testing dataset is selected as the MLKD-Track for inference.

\Remark
By using these different loss functions, the student model can approximate the teacher model in three different ways: through specific numerical distance, internal pixel-wise numerical variation, and approximate range of values. 

\subsection{Comparison with Other Models}


\subsubsection{Comparison with SOTA Tracking Methods}
As shown in Fig.~\ref{fig:rainbow}, MLKD-Track performs better than the SOTA trackers, \textit{i.e.}, LightTrack~\cite{yan2021lighttrack}, SiamFC++~\cite{xu2020siamfc++}, SiamDW~\cite{zhang2019deeper}, SiamRPN++~\cite{siamrpn++}, DaSiamRPN~\cite{zhu2018distractor}, HiFT~\cite{hift}, UpdateNet~\cite{li2021learning}, Ocean~\cite{zhang2020ocean}, SiamAPN~\cite{siamapn}, SiamAPN++~\cite{siamapn++}, and SESiamFC~\cite{sosnovik2021scale}. 
Surpassing the second-best model by \textbf{14.4\%} in success rate and \textbf{18.3\%} in precision, the tracking effectiveness of MLKD-Track can be verified.
MLKD-Track acquires knowledge mainly from the teacher and mutual learning.
Therefore, the effectiveness of the MLKD framework can be  validated.

\subsubsection{Attribute-Based Performance}
To analyze the robustness of MLKD-Track under different challenges, the attribute-based evaluation result on UAVDark135~\cite{adtrack} are shown in TABLE~\ref{tab:II}. The five attributes in nighttime UAV tracking challenges are analyzed, \textit{i.e.}, fast motion, illumination variation, low resolution, occlusion, and viewpoint change challenges.
As shown in TABLE~\ref{tab:II}, MLKD-Track performs the best in all five attributes compared to the other 13 SOTA trackers, surpassing the second-best tracker by over 20\% in many attributes, which can verify the robustness of MLKD-Track.

\subsection{Ablation Study}
In the training process, the student models are trained with two nighttime datasets.
In order to testify to the superior effect of the correlation losses in the training phase, a student model is trained without correlation loss function.
This model is designated as the Student without $\mathcal{L}_{\mathrm{CRL}}$.
Therefore, this student model can only learn from the teacher model at the tracker head.
Meanwhile, the student models without mutual learning are trained respectively with different correlation loss functions.
As shown in TABLE~\ref{tab:ablation}, training in the same condition, the MLKD-Track still has a more precise outcome compared with the other models, demonstrating the significant role of the mutual learning and correlation loss functions.
The correlation losses can guide the training process and result in a more accurate model compared with the trained student model without correlation losses.

The MLKD-Track demonstrates comparable performance to the teacher model and even surpasses it in certain aspects.
Therefore, MLKD-Track is obtained by sharing the knowledge learned by different students.
The student strengths are mutually complemented, achieving comprehensive learning from the teacher.

Furthermore, the speed of all the teacher and student models is tested on a laptop carrying an NVIDIA RTX3060 GPU.
All the student models have the same speed as they have the same parameter quantity.
The speed of the MLKD-Track is shown as the representative of all the student models in TABLE~\ref{tab:ablation}.
MLKD-Track can operate at a higher speed compared with the teacher model, which verified the practicality of MLKD.

These experiments can also testify to the contribution of the mutual-learning knowledge distillation framework.
Thus, the effectiveness of this framework is confirmed.

\begin{figure*}[t]
\centering
\includegraphics[width=1.0\linewidth,scale=1.00]{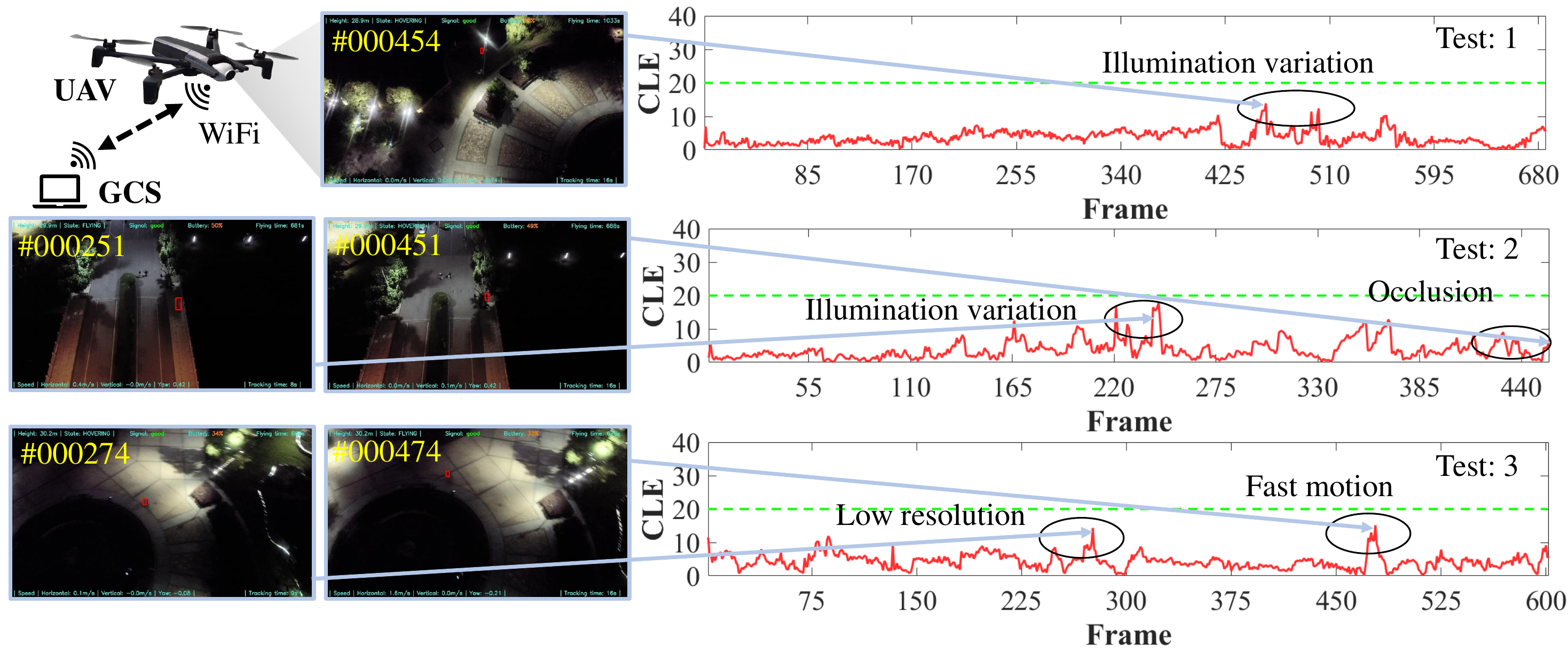}
\vspace{2pt}
\caption{The real-world low-light UAV tracking results and the performance in terms of the CLE. 
The \textcolor[rgb]{1,0,0}{red} lines denote the performance of MLKD-Track. All the tests are regarded as successful with the CLE below the threshold (the \textcolor[rgb]{0,1,0}{green} lines), which validates the practicality of MLKD-Track.
}
\label{fig:real-world}
\end{figure*}




\subsection{Real-World Tests}
The proposed framework is implemented on numerous real-world tests to verify the practicability.
A laptop carrying an NVIDIA RTX3060 GPU serves as the ground control station (GCS).
The Parrot UAV \footnote{https://www.parrot.com/} captures the frames and transmits them to the GCS through WiFi communication.
The tracker predicts the bounding box in real-time, sending it to the Parrot UAV, helping the UAV automatic control.
Exhaustive experiments are conducted to verify the robustness of the MLKD-Track.
The performance of three typical scenes is demonstrated in Fig.~\ref{fig:real-world}.
The CLE curves represent the error between the estimated location and ground truth.
In the tests, MLKD-Track is confronted with fast motion, low resolution, occlusion, and illumination variation challenges.
Based on MLKD, tracking of the UAV is robust and in real-time. 

\begin{table}[!b]
  \centering
  \setlength{\tabcolsep}{9pt}
  \caption{The comparison of all the teacher and student models. The best result is in \textcolor[rgb]{1,0,0}{red} while the second best result is in \textcolor[rgb]{ 0,  .69,  .314}{green}, the MLKD-Track has an advantage against the other student models, confirming the effect of MLKD. Succ represents the success rate, Norm represents the Normalized precision, and Prec represents the precision. ML represents mutual learning, and w/o means without.} 
    \begin{tabular}{c|ccccc}
    \toprule   
    \multicolumn{2}{c}{\textbf{Tracker name}} & \textbf{Succ} & \textbf{Norm} & \multicolumn{1}{c}{\textbf{Prec}} & \multicolumn{1}{c}{\textbf{Speed}} \\
    \midrule
    \multicolumn{2}{c}{Teacher model} &\textcolor[rgb]{ 0,  .69,  .314} {0.461}  & 0.546  &\textcolor[rgb]{ 0,  .69,  .314} {0.601} & 92.3fps \\
    \midrule
    \multicolumn{2}{c}{Student w/o $\mathcal{L}_{\mathrm{CRL}}$}  & 0.450  & 0.534  & 0.576 & \multirow{5}[0]{*}{148.9fps} \\
    \cmidrule{1-5}
    \multirow{3}[0]{*}{w/o ML} & Student 1 & 0.457  & 0.545  & 0.592  \\
    & Student 2 & 0.459 &\textcolor[rgb]{ 0,  .69,  .314}{0.548} & 0.598 & \\
    & Student 3 & 0.446  & 0.535  & 0.579 & \\
    \cmidrule{1-5}
    \multicolumn{2}{c}{\textbf{MLKD-Track}} & \textcolor[rgb]{ 1,  0,  0}{0.468} & \textcolor[rgb]{ 1,  0,  0}{0.558} & \textcolor[rgb]{ 1,  0,  0}{0.602} & \\
    \bottomrule
    \end{tabular}%
  \label{tab:ablation}%
\end{table}%

\section{Conclusions}
This work constructs a mutual-learning knowledge distillation framework for nighttime UAV tracking. 
Concretely, the teacher based on a SOTA enhancer and an advanced baseline tracker backbone is constructed to guide the tight coupling-aware student based on only the advanced baseline tracker backbone in directly extracting the nighttime object features.
Diverse distillation methods are designed to learn different lightweight students focusing on different emphases.
The novel mutual-learning room is proposed to select the best student in each frame to assist the other students.
MLKD-Track can surpass the teacher model in some specific aspects.
Meanwhile, MLKD-Track has a more rapid speed compared to the teacher model.
Furthermore, MLKD-Track performs better than many SOTA trackers.
These experiments verify the effectiveness of MLKD.
Therefore, this work is useful to fasten the model speed while remaining its superior tracking performance and it can achieve the tight coupling between the SOTA enhancer and daytime UAV tracking approach.

\bibliographystyle{IEEEtran}
\balance
\bibliography{reference}

\end{document}